\documentclass[twoside]{article}

\usepackage{hyperref}
\usepackage[dvipsnames]{xcolor}
\usepackage{bm}
\usepackage{amsfonts} 
\usepackage{amsmath}
\usepackage{graphicx}
\usepackage{amsthm}

\usepackage[accepted]{aistats2024}
\usepackage{natbib}
\setcitestyle{authoryear} 

\newtheorem{theorem}{Theorem}[section]

\newtheorem{definition}[theorem]{Definition}

\newtheorem{remark}[theorem]{Remark}

\AtBeginDocument{

  \DeclareMathOperator{\sm}{softmax}

}

\begin{document}

%

%

\twocolumn[

\aistatstitle{$p$-Laplacian Transformer}

\aistatsauthor{ Tuan Nguyen \And Tam Nguyen \And Vinh Nguyen \And  Tan M. Nguyen }

\aistatsaddress{ Toyo University\\Tokyo, Japan \And  Rice University\\Houston, USA \And 
FPT AI Center\\Vietnam \And National University of Singapore\\Singapore } ]

\begin{abstract}
$p$-Laplacian regularization, rooted in graph and image signal processing, introduces a parameter $p$ to control the regularization effect on these data. Smaller values of $p$ promote sparsity and interpretability, while larger values encourage smoother solutions. In this paper, we first show that the self-attention mechanism obtains the minimal Laplacian regularization ($p=2$) and encourages the smoothness in the architecture. However, the smoothness is not suitable for the heterophilic structure of self-attention in transformers where attention weights between tokens that are in close proximity and non-close ones are assigned indistinguishably. From that insight, we then propose a novel class of transformers, namely the $p$-Laplacian Transformer (p-LaT), which leverages $p$-Laplacian regularization framework to harness the heterophilic features within self-attention layers. In particular, low $p$ values will effectively assign higher attention weights to tokens that are in close proximity to the current token being processed. We empirically demonstrate the advantages of p-LaT over the baseline transformers on a wide range of benchmark datasets.
\end{abstract}

\section{Introduction}


In recent years, Transformer models have risen to prominence in the field of machine learning, achieving remarkable success in various domains, including natural language processing \citep{brown2020language,dai2019transformer,raffel2020exploring}, reinforcement learning \citep{chen2021decision}, computer vision \citep{dosovitskiy2020image,liu2021swin,arnab2021vivit,guo2021pct}, and various practical applications \citep{rives2021biological,jumper2021highly,gulati2020conformer,zhang2019deep}. What distinguishes Transformers is their proficiency in transferring knowledge from pre-trained models to new tasks, even when data is limited \citep{radford2018improving,radford2019language,devlin2018bert,liu2019roberta}.

At the core of Transformers lies the self-attention mechanism, a fundamental component that calculates weighted averages of token representations within a sequence. These weightings are determined by assessing similarity scores between pairs of tokens, thereby establishing their relative significance in the sequence. This intrinsic adaptability in capturing a wide array of syntactic and semantic relationships has been identified as a pivotal factor contributing to the triumph of Transformers \citep{vaswani2017attention,vig2019analyzing,voita2019analyzing}.

\subsection {Self-attention Mechanism}
Given an input sequence denoted as $\mathbf{X} := [\boldsymbol{x}(1), \ldots, \boldsymbol{x}(N)]^{\top} \in \mathbb{R}^{N \times D_x}$, self-attention operates by initially projecting $\mathbf{X}$ into three distinct matrices: the query matrix $\mathbf{Q} := [\boldsymbol{q}(1), \ldots, \boldsymbol{q}(N)]^{\top}$, the key matrix $\mathbf{K} := [\boldsymbol{k}(1), \ldots, \boldsymbol{k}(N)]^{\top}$, and the value matrix $\mathbf{V} := [\boldsymbol{v}(1), \ldots, \boldsymbol{v}(N)]^{\top}$. These transformations are achieved through linear operations as described by the following equations:

\begin{equation}
\mathbf{Q} = \mathbf{X} \mathbf{W}_Q^{\top} ; \mathbf{K} = \mathbf{X} \mathbf{W}_K^{\top} ; \mathbf{V} = \mathbf{X} \mathbf{W}_V^{\top}
\label{eq:QKV}
\end{equation}

Here, $\mathbf{W}_Q$ and $\mathbf{W}_K \in \mathbb{R}^{D{qk} \times D_x}$, while $\mathbf{W}_V \in \mathbb{R}^{D \times D_x}$.

Each query vector $\boldsymbol{q}(i)$ corresponds to a specific token at the $i$-th position, and the attention scores from this token to all other tokens are computed by taking the dot product $\boldsymbol{q}(i)^\top\boldsymbol{k}$ between the query vector $\boldsymbol{q}(i)$ and the key vectors $\boldsymbol{v}(j)$. Subsequently, the output vector $\boldsymbol{u}(i)$ is determined by a weighted sum according to the formula:

\begin{equation}\label{eq:self-attention}
\boldsymbol{u}(i) = \sum_{j=1}^N \operatorname{softmax}\left(\boldsymbol{q}(i)^{\top}\boldsymbol{k}(j)/\sqrt{D_{q k}}\right) \boldsymbol{v}(j)
\end{equation}

Alternatively, the output sequence $\mathbf{U} := [\boldsymbol{u}(1), \ldots, \boldsymbol{u}(N)]^{\top} \in \mathbb{R}^{N \times D_{qk}}$ can be computed as:

\begin{equation*}
\mathbf{U} = \operatorname{softmax}\left(\mathbf{Q} \mathbf{K}^{\top}/\sqrt{D_{qk}}\right) \mathbf{V} := \mathbf{A} \mathbf{V}
\end{equation*}

Here, the softmax function is applied row-wise. The matrix $\mathbf{A}$, denoted as the softmax attention matrix, belongs to $\mathbb{R}^{N \times N}$ and its elements represent the computed attention scores.

\subsection{Multi-Head Self-Attention}
The Multi-Head Self-Attention (MSA) mechanism plays a pivotal role in the Transformer architecture, a core component that empowers these models with their remarkable capabilities. MSA leverages a group of self-attention heads, seamlessly combining their outputs through a linear projection \citep{vaswani2017attention}
\begin{equation*}
    \text{MSA}(\mathbf{X}):=\text{FFN}\big([\mathbf{U}^1,\dots,\mathbf{U}^H]\big)
\end{equation*}
Here, $H$ represents the total number of attention heads, and $\mathbf{U}^h$ represents the output of the self-attention mechanism at the $h$-th head. Each head is equipped with its parameter sets ${\mathbf{W}_Q^h;\mathbf{W}_K^h;\mathbf{W}_V^h}$, corresponding to the query, key, and value matrices $\{\mathbf{Q}^h;\mathbf{K}^h;\mathbf{V}^h\}$ as defined in Equation \eqref{eq:QKV}.

The final linear projection, denoted as $\text{FFN}$, can be expressed as:
\begin{equation*}
    \text{FFN}(\mathbf{X}):=\mathbf{X}\mathbf{W}_O
\end{equation*}

Here, $\mathbf{W}O \in \mathbb{R}^{HD{qk} \times D_{qk}}$ serves as the essential operator that projects the multi-head outputs into the corresponding hidden dimensions. Besides MSA module, each transformer block is equipped with a skip connections to cooperate with MSA. Formally, a transformer layer $l$-th can be written as follows:
\begin{equation*}
    \mathbf{U}_{(l+1)}=\text{MSA}(\mathbf{U}_l) + \mathbf{U}_l
\end{equation*}

\subsection{Transformers and Heterophily structures}


Heterophily, which is often associated with a disassortative structure, characterizes networks where connected nodes tend to belong to different classes or exhibit dissimilar attributes. It's a concept deeply rooted in network science, reflecting the propensity of diverse elements to connect. Heterophily can manifest in a variety of domains, for instance, in a network based on word adjacencies in text, where the labels represent the part of speech (POS) of each word, it's common to observe that adjacent words frequently have distinct POS labels \citep{gu2023enhancing}. Words, even when closely positioned within the text, often serve different linguistic functions, leading to this inherent diversity in their attributes. Another notable example can be found in the realm of protein chemistry, where chemical interactions often occur between different types of amino acids \citep{li2022finding}. However, it's important to note that the current self-attention, primarily designed for tasks involving homophilic structures, is not suitable for effectively modeling heterophily relationships.


In general, Self Attention operates by updating token representations through the aggregation of information from other tokens, effectively acting as a kind of low-pass filter \citep{wang2022anti}. This filter primarily emphasizes the commonalities in token features while inadvertently downplaying their differences, resulting in a homogenization of learned token representations. The inherent smoothness of low-frequency information makes this mechanism well-suited for networks exhibiting homophily where connected nodes tend to belong to the same classes or similar attributes. However, for heterophilic networks, applying a smoothing filter across connections may not be ideal, as it encourages connected nodes to resemble each other predictably, contradicting the nature of disassortativity. For instance, imposing similar representations on connected proteins through a low-pass filter would significantly impede performance, as the low-frequency information struggles to support inferences within such networks. In such cases, high-frequency information, which captures the nuanced differences between nodes, may offer a more appropriate solution.


\subsection{Contribution}

Under heterophily, it is crucial to leverage both similarities and dissimilarity between the node features in a neighborhood to be able to learn high-quality representations. However, without having access to the labels, it is not clear how the node features in a neighborhood should be aggregated. To address the above challenge, our key idea is to leverage a low-pass filter in a typical Transformers encoder to capture both similarities and dissimilarities of nodes with their neighborhood and generate representations.

One potential approach to address this issue is through the nonlocal framework. In nonlocal framework \citep{gilboa2009nonlocal}, image patches are treated as nodes in a weighted graph, with each edge assigned a value that quantifies the similarity between the corresponding nodes. We can formally formulate the relationships between patches and their collective role within the entire image. This enables a more thorough study of the intricacies of images. Additionally, this approach allows for a convenient definition of energy functionals and can be further put into PDE-like evolutions for comprehensive processing and analysis \citep{rudin1992nonlinear}. Furthermore, it has been shown that self-attention is closely related to the nonlocal framework \citep{wang2018non,ramachandran2019stand}. In this study, we delve into the $p$-Laplacian regularization framework as a fundamental component of our exploration, to understand its implications for the self-attention mechanism and performance of Transformers.

Our main contributions can be summarized as follows
\begin{enumerate}
    \item We use $p$-Laplacian Regularization Framework to show that the self-attention mechanism obtains the minimal $p$-laplacian regularized functional in the case of $p=2$.
    \item Motivated from $p$-Laplacian Regularization Framework, we propose $p$-Laplacian Transformers ($p$-LaT).
    \item We theoretically prove that $p$-LaT guarantees the convergence of $p$-LaT and can alleviate the low-pass filter effect.
\end{enumerate}

We empirically demonstrate the advantages of $p$-LaT on various large-scale applications, including the ImageNet object classification, and WikiText-103 language modeling tasks.

\textbf{Organization}. Our paper follows this structured path: In Section \ref{sec:p=2}, we reveal the intrinsic link between self-attention and the $p$-Laplacian regularization framework, particularly when $p=2$. Section \ref{sec:p-lat} introduces the novel $p$-LaT model, underpinned by theoretical guarantees of convergence. A comprehensive spectral analysis in Section \ref{sec:spectral} showcases the non-low-pass filters characteristic in $p$-LaT. Section \ref{sec:experiment} is dedicated to empirical validation across various benchmarks. We provide context within the realm of related work in Section \ref{sec:related}, and in our concluding section, we summarize our contributions and offer broader insights.

\section{Self-attention is a particular case of $p$-regularization}
\label{sec:p=2}

In this section, we establish a strong connection between traditional self-attention mechanisms and the $p$-Laplacian regularization framework. Specifically, we derive Equation \eqref{eq:self-attention} from the $p$-Laplacian regularization framework, in the case where $p=2$. We then reveal that a layer of self-attention can be viewed as a step in the minimization process of a functional. This connection not only enhances our understanding of self-attention but also underscores the adaptability of the $p$-Laplacian regularization framework beyond its conventional domain in image processing.



We first consider the output matrix $ \mathbf{U} := [\boldsymbol{u}(1), \ldots, \boldsymbol{u}(N)]^{\top} \in \mathbb{R}^{N \times D}$ in \eqref{eq:self-attention} be a real vector-valued function, i.e., $\bm{u}:\Omega \rightarrow \mathbb{R}^D$, $u \in C^2(\Omega)$ with $\Omega \in \mathbb{R}$ and $x \in \Omega$. Then we can present the process of minimizing the following $p$-laplacian regularized functional
\begin{align}
    J(\bm{u})&=\frac{1}{p}\int_{\Omega^2}k(x,y)\left\|\bm{u}(y)-\bm{u}(x)\right\|^p dydx
    \label{functional}
\end{align}

To minimize the functional $J(\bm{u})$, we process by assuming that there is $\bm{u} \in C^2(\bar{\Omega})$ so that $J(\bm{u}) \leq J(\bm{v})$ for all $v \in C^2(\bar{\Omega})$ and $\bm{u}=\bm{v}$ on $\partial\Omega$. Now, let $h \in C^\infty_{c}(\Omega)$ and set $\bm{v}=\bm{u}+t\bm{h}$ for a real number $t$. Thus, by the above assumption we have
\begin{equation*}
    J(\bm{u}) \leq J(\bm{v})=J(\bm{u}+t\bm{h}) \quad \forall t \in \mathbb{R}
\end{equation*}
This indicates that $J(\bm{u}+t\bm{h})$ has a global minimum at $t=0$. Now we need to solve the following equation to minimize \eqref{functional}
\begin{equation}
    {\frac{d}{dt}}\biggr\rvert_{t=0} J(\bm{u}+t\bm{h})=0
    \label{g_0}
\end{equation}
For simplicity, we define $I: \mathbb{R}^D \rightarrow \mathbb{R}, \quad I(Q)=Q^p$.


This yields
\begin{align*}
    \frac{d}{dt} J(\bm{u}+t\bm{h}) &= \frac{1}{p} \int_{\Omega^2} {k(x,y)\frac{d}{dt} I\biggr(\bm{u}(x)-\bm{u}(y)}\\
    &\quad\quad\quad\quad\quad\quad\quad\quad+ t\big(h(x)-h(y)\big)\biggr) dxdy\\
    \frac{d}{dt} J(\bm{u}+t\bm{h}) &= \frac{1}{p} \int_{\Omega^2}{k(x,y)\nabla I\biggr(\bm{u}(x)-\bm{u}(y)}\\
    &\quad\quad+ t\big(\bm{h}(x)-\bm{h}(y)\big)\biggr)\big(\bm{h}(x)-\bm{h}(y)\big) dxdy
\end{align*}
Therefore, from \eqref{g_0} we have
\begin{align*}
    {\frac{d}{dt}}\biggr\rvert_{t=0}J(\bm{u}+t\bm{h}) =\frac{1}{p} \int_{\Omega^2}{k(x,y)\nabla I\biggr(\bm{u}(x)-\bm{u}(y)\biggr)}\\
    \biggr(\bm{h}(x)-\bm{h}(y)\biggr) dxdy =  0  \\
\end{align*}
Or,
\begin{align*}
     \frac{1}{p} &\int_{\Omega^2}{k(x,y)\nabla I\biggr(\bm{u}(x)-\bm{u}(y)\biggr)\bm{h}(x) dxdy} 
     \\- \frac{1}{p} &\int_{\Omega^2}{k(x,y)\nabla I\biggr(\bm{u}(x)-\bm{u}(y)\biggr)\bm{h}(y) dxdy} = 0
\end{align*}

and change of variable $(x,y) \to (y,x)$ in the second integral and with
\begin{equation*}
    \nabla I\big(\bm{u}(x)-\bm{u}(y)\big)=p\|\bm{u}(x)-\bm{u}(y)\|^{p-2}\big(\bm{u}(x)-\bm{u}(y)\big)
\end{equation*} 
we obtain
\begin{equation}
    \int_{\Omega^2}{K(x,y)\|\bm{u}(x)-\bm{u}(y)\|^{p-2}\big(\bm{u}(x)-\bm{u}(y)\big)\bm{h}(x) dxdy}= 0
    \label{euler_h}
\end{equation}
where $K(x,y):=k(x,y)+k(y,x)$ is a symmetric kernel. 

Since \eqref{euler_h} is true for all $h \in C^\infty_c(\Omega)$, we deduce
\begin{equation}
      \int_{\Omega}{K(x,y)\|\bm{u}(x)-\bm{u}(y)\|^{p-2}(\bm{u}(x)-\bm{u}(y)) dy}= 0
      \label{euler}
\end{equation}

To numerically compute solutions of \eqref{euler}, one could use the gradient descent method, which corresponds to solving the PDE

\begin{align}
    \frac{d\bm{u}(x,t)}{dt}&=-\int_{\Omega}{K(x,y)\bm{p}(x,y)(\bm{u}(x)-\bm{u}(y)) dy} \nonumber\\
    \bm{p}(x,y)&=\|\bm{u}(x)-\bm{u}(y)\|^{p-2}
    \label{grad_flow}
\end{align}


We discretize the ODE in \eqref{grad_flow} using Euler method with step size $\Delta t = 1/\int_\Omega K(x,y) dy$ and obtaining the following

\begin{align*}
    &\bm{u}(x,\Delta t) = \bm{u}(x,0)+
    \\&\Delta t \int_{\Omega}{K(x,y)\|\bm{u}(x,0)-\bm{u}(y,0)\|^{p-2}\biggr(\bm{u}(y,0)-\bm{u}(x,0)\biggr) dy}\\
    &= \int_{\Omega}{\frac{K(x,y)\|\bm{u}(x,0)-\bm{u}(y,0)\|^{p-2}}{\int_{\Omega}{K(x,y^\prime) dy^\prime}}\bm{u}(y,0) dy}\\
    &= \int_{\Omega}{\frac{K(x,y)}{\int_{\Omega}{K(x,y^\prime) dy^\prime}} \bm{u}(y,0) dy}
\end{align*}
the last line is obtained by setting $p=2$. 

Let $\bm{v}(x):=[v_1(x),\dots,v_D(x)]^\top$ be a real vector-valued function, where $\bm{v}:\Omega \rightarrow \mathbb{R}^D$, $\bm{v} \in C^2(\Omega)$. We set initial value of $\bm{u}$ with $\bm{v}(x)$, i.e., $\bm{u}(x,0)=\bm{v}(x)$. Similar to $\bm{v}(x)$, we define $\bm{k}(x):=[k_1(x),\dots,k_{D_{qk}}(x)]^\top$ be a real vector-valued function, where $\bm{k}:\Omega \rightarrow \mathbb{R}^{D_{qk}}$, $\bm{k} \in C^2(\Omega)$. We choose $K(x,y)=\exp(\bm{k}(x)^\top\bm{k}(y)/\sqrt{D_{qk}})$ and using Monte-Carlo approximation, we could rewrite the above equation as following
\begin{align}
    \bm{u}(x,\Delta t) &\approx \sum_{y=1}^N{\frac{\exp(\bm{k}(x)^\top\bm{k}(y)/\sqrt{D_{qk}})}{\sum_{y^\prime=1}^N{\exp(\bm{k}(x)^\top\bm{k}(y^\prime)/\sqrt{D_{qk}})}} \bm{v}(y)} \nonumber\\
     &\approx \sum_{y=1}^N\text{softmax}\left(\bm{k}(x)^\top\bm{k}(y) / \sqrt{D_{qk}}\right)\bm{v}(y)
    \label{discrete_attn}
\end{align}

Comparing \eqref{discrete_attn} and \eqref{eq:self-attention}, we observe that \eqref{discrete_attn} implement a symmetric self-attention, in which the query matrix $\mathbf{Q}$ and the key matrix $\mathbf{K}$ are identical, i.e. $\mathbf{W}_Q = \mathbf{W}_K$ where $\mathbf{W}_Q$ and $\mathbf{W}_K$ are the linear projections that map the input sequence $\mathbf{X}$ into $\mathbf{Q}$ and $\mathbf{K}$ as given in \eqref{eq:QKV}. If we relax this symmetry condition, in other words, we replace the key vectors $\bm{k}(x)$ with the query vectors $\bm{q}(x), x = 1,\dots, N$, to obtain
the exact formula of self-attention given by \eqref{eq:self-attention} would be obtained. The following theorem summarizes our above result

\begin{theorem}
    Given the $p$-laplacian regularized functional $J(\bm{u})=\frac{1}{p}\int_{\Omega^2}k(x,y)\left\|\bm{u}(y)-\bm{u}(x)\right\|^p dydx$ of a vector-valued function $\bm{u}:\Omega \rightarrow \mathbb{R}^D$, $\bm{u} \in C^2(\Omega)$. The corresponding gradient flow to minimize $J(\bm{u})$ is $\frac{d\bm{u}(x,t)}{dt}=-\int_{\Omega}{K(x,y)\|\bm{u}(x)-\bm{u}(y)\|^{p-2}(\bm{u}(x)-\bm{u}(y)) dy}$ where $K(x,y)=\exp(\bm{k}(x)^\top\bm{k}(y)/\sqrt{D_{qk}})$ and $\bm{k}:\Omega \rightarrow \mathbb{R}^D$, $\bm{k} \in C^2(\Omega)$. Then, taking an Euler discretization method with initial value $\bm{u}(x,0)=\bm{v}(x)$ and step size $\Delta t =  1/\int_\Omega K(x,y) dy$ is equivalent to a symmetric self-attention
    \begin{equation}
            \bm{u}(x) = \sum_{y=1}^N\sm\left(\bm{k}(x)^\top\bm{k}(y) / \sqrt{D_{qk}}\right)\bm{v}(y)
            \label{theorem:sym_att}
    \end{equation}
    Breaking the symmetry of the attention scores by replacing $\bm{k}(x)$ with $\bm{q}(x)$, $x = 1,\dots, N$, in \eqref{theorem:sym_att}, we obtain the exact formula of self-attention
    \begin{equation}
            \bm{u}(x) = \sum_{y=1}^N\sm\left(\bm{q}(x)^\top\bm{k}(y) / \sqrt{D_{qk}}\right)\bm{v}(y)
            \label{theorem:att0}
    \end{equation}

\end{theorem}

\section{$p$-Laplacian Transformers}
\label{sec:p-lat}

In this section, we derive the $p$-Laplacian-based self-attention equation from a $p$-Laplacian regularization framework and present $p$-LaT, a new Transformers architecture developed upon the new self-attention scheme. We then provide a theoretical result that the new self-attention scheme is guaranteed to converge.

Following the derivations from section 2, from ODE \eqref{grad_flow}, we again discretize using the Euler method, but this time, with arbitrary $p$ value and  step size $\Delta t = 1/\int_\Omega K(x,y) dy$, initial value $\bm{u}(x,0)=\bm{v}(x)$

\begin{align*}
    \bm{u}(x,\Delta t) &= \int_{\Omega}{\frac{K(x,y)}{\int_{\Omega}{K(x,y^\prime) dy^\prime}}\|\bm{v}(x)-\bm{v}(y)\|^{p-2}\bm{v}(y) dy}
\end{align*}

Then choose $K(x,y)=\exp(\bm{k}(x)^\top\bm{k}(y)/\sqrt{D_{qk}})$ and using the Monte-Carlo method, we would obtain the following new formula for calculating symmetric self-attention

\begin{align*}
    \bm{u}(&x,\Delta t) \approx
    \\&\sum_{y=1}^N\text{softmax}\left(\frac{\bm{k}(x)^\top\bm{k}(y)}{\sqrt{D_{qk}}}\right)\|\bm{v}(x)-\bm{v}(y)\|^{p-2}\bm{v}(y)
    \label{p_lap_sym_attn}
\end{align*}

 By replacing the key vectors $\bm{k}(x)$ with the query vectors $\bm{q}(x)$, $x = 1, \dots, N$, its corresponding asymmetric self-attention is

\begin{equation}
        \bm{u}(x) = \sum_{y=1}^N\text{softmax}\left(\frac{\bm{q}(x)^\top\bm{k}(y)}{\sqrt{D_{qk}}}\right)\|\bm{v}(x)-\bm{v}(y)\|^{p-2}\bm{v}(y)
        \label{theorem:att}
\end{equation}

With \eqref{theorem:att}, we are now ready to define the $p$-Laplacian Transformers ($p$-LaT).
\begin{definition}[$p$-Laplacian Transformers]
    For each layer $l=1,\dots,L$ of $p$-LaT, the set of query, key and value vectors $\{\bm{q}^l(x),\bm{k}^l(x),\bm{v}^l(x)\}_{x=1}^N$ of output sequence $ \mathbf{U}^{l-1} := [\boldsymbol{u}^{l-1}(1), \ldots, \boldsymbol{u}^{l-1}(N)]^{\top} \in \mathbb{R}^{N \times D}$ in the previous layer will be set by formulae in \eqref{eq:QKV}. And the corresponding output vector $\bm{u}^l(x)$, $x=1,\dots,N$ of layer $l$-th will be computed by the following attention formula with arbitrary $p$ value
    \begin{align}
        \bm{u}^l(x) &= \sum_{y=1}^N\sm\left(\frac{\bm{q}^{l}(x)^\top\bm{k}^l(y)}{\sqrt{D_{qk}}}\right)\bm{p}^l(x,y)\bm{v}^l(y) \nonumber\\
        \bm{p}^l(x,y)&=\|\bm{v}^l(x)-\bm{v}^l(y)\|^{p-2}
        \label{p-lat}
    \end{align}
\end{definition}

\begin{remark}
    $\mathbf{U}^0$ sequence is identical to the input sequence $\mathbf{X} := [\boldsymbol{x}(1), \ldots, \boldsymbol{x}(N)]^{\top} \in \mathbb{R}^{N \times D_x}$ of the Transformers.
\end{remark}

\begin{remark}
    Rewrite \eqref{p-lat} in a matrix form we obtain $\mathbf{U}_{l}= \operatorname{softmax}\left(\mathbf{Q}_l \mathbf{K}_l^{\top}/\sqrt{D_{qk}}\right) \mathbf{P}_l\mathbf{V}_l$ where $\mathbf{P}_l \in \mathbb{R}^{N \times N}$ and the position $(x,y)$ in that matrix has the value of $\mathbf{P}_l(x,y)=\|\bm{v}^l(x)-\bm{v}^l(y)\|^{p-2}$.
\end{remark}

We present the following theorem to show the convergence property of $p$-LaT.

\begin{theorem}
    \label{theorem:convergence}
    Given sequence $ \mathbf{U}^{l} := [\boldsymbol{u}^{l}(1), \ldots, \boldsymbol{u}^{l}(N)]^{\top} \in \mathbb{R}^{N \times D}$ is the output of $p$-LaT at layer $l$-th and the $J$ defined in \eqref{functional}. Then
    \begin{equation}
        J(\mathbf{U}^l) \geq J(\mathbf{U}^{l+1})
    \end{equation}
\end{theorem}

Theorem \ref{theorem:convergence} shows that the functional in \eqref{functional}, or the objective function, is guaranteed to decline after
taking one layer of $p$-LaT.

\begin{proof}
    To prove the theorem, we need to show that the right-hand side of \eqref{grad_flow} is the direction of steepest descent and minimizes $J(\bm{u})$. Again, for notational simplicity, we define $L(x)=\int_{\Omega}{K(x,y)\|\bm{u}(x)-\bm{u}(y)\|^{p-2}(\bm{u}(x)-\bm{u}(y)) dy}$. From the left-hand-side of \eqref{euler_h}, we recall that
    \begin{align*}
       &{\frac{d}{dt}}\biggr\rvert_{t=0}J(\bm{u}+t\bm{h})= 
       \\&\int_{\Omega^2}{K(x,y)\|\bm{u}(x)-\bm{u}(y)\|^{p-2}}\biggr(\bm{u}(x)-\bm{u}(y)\biggr)\bm{h}(x) dxdy\\
        &= \biggr<\int_{\Omega}{K(x,y)\|\bm{u}(x)-\bm{u}(y)\|^{p-2}}\biggr(\bm{u}(x)-\bm{u}(y)\biggr) dy,\bm{h}(x)\biggr>_{L^2(\Omega)}\\
        &= \big<L(x), \bm{h}(x)\big>_{L^2(\Omega)}
    \end{align*}
    Then using the Cauchy-Schwarz inequality,
    \begin{equation*}
        \biggr| {\frac{d}{dt}}\big\rvert_{t=0}J(\bm{u}+t\bm{h}) \biggr| \leq \|L(x)\|\cdot\|\bm{h}(x)\|
    \end{equation*}
    the equality happens iff $\bm{h}(x) = \alpha L(x)$ with a positive number $\alpha$.  This implies that the direction $\bm{h}(x) = \alpha L(x)$ is the direction of the steepest ascent and maximizes $J(\bm{u})$. And thus $-L(x)=-\int_{\Omega}{K(x,y)\|\bm{u}(x)-\bm{u}(y)\|^{p-2}(\bm{u}(x)-\bm{u}(y)) dy}$ points in the direction of steepest descent, and the flow in \eqref{grad_flow} tends to move toward a state of lesser energy.
\end{proof}



\section{Spectral Analysis of $p$-LaT}
\label{sec:spectral}

In this section, we embark on a spectral analysis journey to unveil the distinct characteristics of $p$-LaT that set them apart from traditional Transformers. Specifically, we focus on demonstrating that \eqref{theorem:att} does not conform to the attributes of a conventional low-pass filter.

Our analytical tool of choice is the Fourier transform, which serves as a powerful instrument in our investigation. To lay the foundation for our spectral analysis, we begin by revisiting the fundamentals of Fourier analysis and the definition of endomorphism low-pass filters, as outlined in \cite{wang2022anti}.

Denote $\mathcal{F}:\mathbb{R}^n \rightarrow \mathbb{C}^n$ be the Discrete Fourier Transform (DFT) with the Inverse Discrete Fourier Transform (IFT) $\mathcal{F}^{-1}:\mathbb{C}^n \rightarrow \mathbb{R}^n$. $\mathbf{f}_k =[e^{2\pi j (k-1) \cdot 0 }\dots e^{2\pi j (k-1) \cdot (n-1) }]^\top / \sqrt{n} \in \mathbb{R}^n$ where $k$ denotes the k-th row of DFT matrix, and $j$ is the imaginary unit.

Let $\Tilde{z}=\mathcal{F}z$ be the spectrum of $z$, and $\Tilde{z}_{dc} \in \mathbb{C}, \Tilde{z}_{hc} \in \mathbb{C}^{n-1}$ take the first element and the rest elements of $\Tilde{z}$, respectively. Define $\mathcal{DC}[z] = \Tilde{z}_{dc}\mathbf{f}_1 \in \mathbb{C}^n$ as the Direct-Current (DC) component of signal z, and $\mathcal{HC}[z] = [\mathbf{f}_2 \dots \mathbf{f}_n] \Tilde{z}_{hc} \in \mathbb{C}^n$ the complementary high-frequency component. A low-pass filter is a system that suppresses the high-frequency component of signals and retains the low-frequency component. Thus, a low-pass filter will only preserve the DC component, while diminishing the remaining HC component.
\begin{figure*}[!ht]
  \centering
    \includegraphics[width=\textwidth]{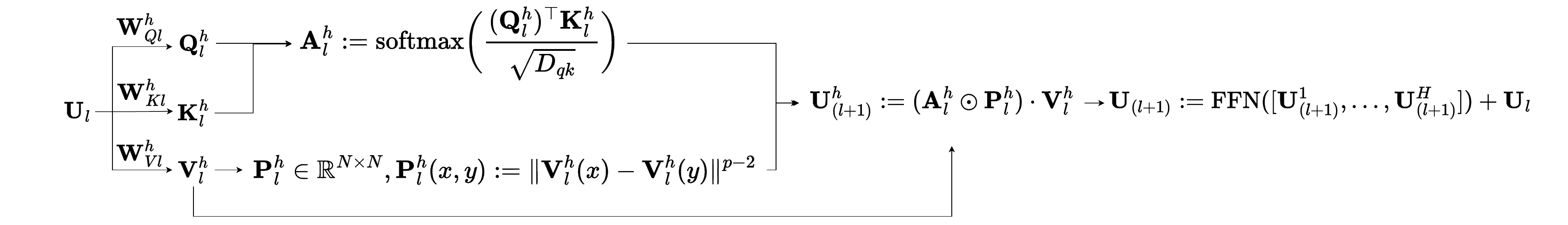}
    \centering
    \caption{Architecture of a $p$-Laplacian Transformer at one head. We denotes $H$ is the number of heads and $h=1,\dots,H$. The set query, key, value matrices at layer $l$-th and head $h$-th are denotes as $\{\mathbf{Q}_l^h;\mathbf{K}_l^h;\mathbf{V}_l^h\}$}. Also, $FFN$ is the feed-forward layer and $[\mathbf{U}_{(l+1)}^1;\dots;\mathbf{U}_{(l+1)}^H]$ is the concatenate operator.
  \label{fig:archi}    
\end{figure*}

\begin{definition}
    Given an endomorphism $f:\mathbb{R}^N \to \mathbb{R}^N$, $f$ is a
low-pass filter if and only if
\begin{eqnarray*}
  \lim_{t \to \infty} \frac{||\mathcal{HC}(f^t z)||
}{||\mathcal{DC}(f^t z)||}=0, \quad \forall z \in \mathbb{R}^N.
\end{eqnarray*}
\end{definition}

The main result of that paper is that self-attention is constantly a low-pass filter. More precisely, they proved the following theorem.

\begin{theorem}[\cite{wang2022anti}]
    Let $\mathbf{A} = softmax(\mathbf{P})$, where $\mathbf{P} \in \mathbb{R}^{n \times n}$. Then $\mathbf{A}$ must be a low-pass filter  for all $z \in \mathbb{R}^n$:
    \begin{equation}
        \lim_{t \rightarrow \infty}\frac{\|\mathcal{HC}[\mathbf{A}^tz]\|}{\|\mathcal{DC}[\mathbf{A}^tz]\|}=0
    \end{equation}
\end{theorem}

Since $\mathbf{A}$ is the output of softmax function, i.e. a right-stochastic matrix, it has the largest eigenvalue $\lambda_{max}=1$. We found that the need for $\mathbf{A}$ to be a right-stochastic matrix is a tight condition. In fact, we can loosen up and make the theorem more general. We state the following theorem

\begin{theorem}
    Let $\mathbf{A}$ be a positive matrix, in which it has $\lambda_{max} \leq 1$. Then $\mathbf{A}$ must be a low-pass filter  for all $z \in \mathbb{R}^n$:
    \begin{equation}
        \lim_{t \rightarrow \infty}\frac{\|\mathcal{HC}[\mathbf{A}^tz]\|}{\|\mathcal{DC}[\mathbf{A}^tz]\|}=0
    \end{equation}
    \label{theorem:general_low_pass}
\end{theorem}

To prove the theorem, we only need to make a slight modification in the proof of the original paper. 


Moreover, we can establish the subsequent theorem, which represents the converse of Theorem \ref{theorem:general_low_pass}. It effectively demonstrates how, through the application of $p$-Laplacian regularization, we can mitigate the effects of low-pass filters. While Theorem \ref{theorem:general_low_pass} may not stand as a central theorem in our research, it forms an indispensable stepping stone, laying the groundwork for the theorem that follows.

\begin{theorem}
    Let $\mathbf{A}$ be a low-pass filter. If $\mathbf{1}$ is not the eigenvector correspond to $\lambda_{max}$, then we have $\lambda_{max} \leq 1$.
    \label{theorem:eigen_p_lap}
\end{theorem}

\begin{proof}
    We now prove the Theorem \ref{theorem:eigen_p_lap} by contradiction. We assume that $\bf{1}$ is not an eigenvector associated with $\lambda_{max}$,  $\lambda_{max} > 1$, and
\begin{eqnarray*}
  \lim_{t \to \infty} \frac{||\mathcal{HC}(A^t z)||
}{||\mathcal{DC}(A^t z)||}=0, \text{for all $z \in \mathbb{R}^N$}.
\end{eqnarray*} 
We will try to deduce a contradiction.

The Perron-Frobenius theorem \citep{meyer2023matrix} claims that there is a positive vector (element-wise) $y$ such that $\mathbf{A}y=\lambda_{max} y$. This implies $\mathbf{A}^ty=\lambda_{max}^t y$. Insert this equality into the above term, we get
\begin{align}
  \lim_{t \to \infty} \frac{||\mathcal{HC}(\mathbf{A}^t y)||
}{||\mathcal{DC}(\mathbf{A}^t y)||}&=\lim_{t \to \infty}\sqrt{\frac{||\mathcal{HC}(\mathbf{A}^t y)||^2}{||\mathcal{DC}(\mathbf{A}^t y)||^2}} \nonumber
\\&=\lim_{t \to \infty}\sqrt{\frac{||\lambda_{max}^t (I-\frac{1}{N}1 1^T)y||^2
}{||y-\lambda_{max}^t (I-\frac{1}{N}1 1^T)y||^2}}
\label{cal}
\end{align}

Since we assume that the vector $y$ is not a multiplication of $\bf{1}$, the last term in~\ref{cal} is equal to 1.
\end{proof}

The significance of Theorem \ref{theorem:eigen_p_lap} lies in its capacity to elucidate how achieving $\lambda_{max} > 1$ can disrupt the low-pass filter characteristics inherent in $\mathbf{A}$. We can attain this condition, for example, when $\sum_j \mathbf{A}_{ij} \geq 1, \forall i$ and there exists at least one row that $\sum_j \mathbf{A}_{ij} > 1$.


Given the unavailability of token labels within the intermediate layers of Transformers, we define homophily and heterophily in terms of features vectors similarity and dissimilarity.

\begin{table*}[t]
\centering
    \caption{Top-1 and Top-5 accuracy (\%) of $p$-LaT vs DeiT on the ImageNet benchmark.}
    {\begin{tabular}{ccc}
         \hline
          Model & Top-1 Acc (\%)& Top-5 Acc (\%) \\  
         \hline 
         DeIT & 71.97 & 91.13\\
         \hline 
         $p$-LaT& \textbf{72.78} & \textbf{91.31}\\
         \hline
    \end{tabular}}
    \label{tab:result_imagenet}
\end{table*}

\begin{table*}[t]
\centering
    \caption{Test and valid perplexity (Test PPL and Valid PPL) on WikiText-103 of $p$-LaT compared to the conventional softmax transformer.}
    {\begin{tabular}{ccc}
         \hline
          Model & Valid PPL&Test PPL \\  
         \hline 
         Softmax Transformer & 33.17 & 34.1\\
         \hline 
         $p$-LaT& \textbf{32.57} & \textbf{33.5}\\
         \hline
    \end{tabular}}
    \label{tab:result_wiki}
\end{table*}
\begin{definition}
For a specific layer, denoted as the $l$-th layer of the Transformers, we characterize the implicit structure as homophily when it satisfies:
\begin{equation*}
\|\bm{v}^l(x) - \bm{v}^l(y)\| < 1, \quad \forall{x,y}
\end{equation*}
In contrast, the implicit structure is classified as heterophily if it meets the condition:
\begin{equation*}
\|\bm{v}^l(x) - \bm{v}^l(y)\| \geq 1, \quad \forall{x,y}
\end{equation*}
    
\end{definition}


The introduction of the $p$-laplacian term, $\|\bm{v}(x) - \bm{v}(y)\|^{p-2}$, leads to a transformation in the eigenvalue characteristics. Specifically, the presence of the $p$-laplacian term results in $\lambda_{max} > 1$ for the $p$-LaT, rendering matrix $\mathbf{A}$ as a non-low-pass filter.

For the case of $p < 2$, the $p$-LaT model escapes from low-pass filters situation for $\|\bm{v}(x) - \bm{v}(y)\| < 1, \forall{x,y}$. This observation highlights the effectiveness of the $p$-LaT model in in the case of strong homophily structure.

Conversely, for the scenario where $p > 2$, the $p$-LaT model no longer acts as a low-pass filter for $\|\bm{v}(x) - \bm{v}(y)\| > 1, \forall{x,y}$. This underscores the $p$-LaT model's proficiency in modeling strong heterophily structure.

\begin{remark}
    In the form of matrix, $\mathbf{A}$ in the case of $p$-LaT is defined as $\mathbf{A}:=\sm(\mathbf{Q}^\top \mathbf{K} / \sqrt{D_{qk}})\mathbf{P}$. Where $\mathbf{P} \in \mathbb{R}^{N \times N}$ and the position (x,y) in that matrix has the value of $\mathbf{P}(x,y):=\|\mathbf{V}(x)-\mathbf{V}(y)\|^{p-2}$. 
\end{remark}


\section{Experimental Results}
\label{sec:experiment}

In this section, we empirically demonstrate the advantages of our proposed $p$-LaT approach across various tasks, including ImageNet classification\citep{deng2009imagenet} and language modeling on the WikiText-103\citep{merity2016pointer}. All experiments were conducted on a single DGX-A100 system.

\textbf{Object classification on ImageNet}. To demonstrate the advantage of our method, we compare it with the DeiT baseline\citep{touvron2021training} on the ImageNet image classification task. The ImageNet dataset\citep{deng2009imagenet} comprises 1.28 million training images and 50, 000 validation images, encompassing the classification of 1000 categories. The evaluation metrics used for performance assessment are the top-1 and top-5 accuracies. 

\textbf{Models settings for Imagenet}. Our baseline model is the DeiT-tiny model\citep{touvron2021training}, which consists of 12 transformer layers, 3 attention heads per layer, and a model dimension of 192. For model setting and setting and configuration, we follow \cite{touvron2021training}. For $p$-LaT, we follow the same parameters settings, the difference is that we set 2 attention heads with low $p$ value ($p=1.5$), and the other 2 heads with high $p$ value ($p=2.5$). Figure \ref{fig:archi} represents our architecture.

Our $p$-LaT outperforms the DeiT baseline, demonstrating superior performance in both Top-1 Accuracy and Top-5 Accuracy metrics, as illustrated in Table \ref{tab:result_imagenet}.

\textbf{Language Model on WikiText-103}. In addition to computer vision tasks, we also evaluate the effectiveness of our model on a large-scale natural language processing application, specifically language
modeling on WikiText-103. The WikiText-103 dataset is a carefully curated collection of articles sourced from Wikipedia, designed to capture intricate and extensive contextual dependencies. The training set comprises a substantial 28,000 articles, totaling an impressive 103 million running words. Each article is structured into text blocks, each spanning around 3,600 words. Moving to the validation and test sets, they encompass 218,000 and 246,000 running words, respectively, each composed of 60 articles totaling approximately 268,000 words. Our experimental framework adheres to standard practices \citep{schlag2021linear,merity2016pointer}, which involve segmenting the training data into independent sequences, each comprising $L$ words. For our evaluation, we employ a batch size of 1 and process the text sequence through a sliding window of size $L$. When assessing perplexity (PPL), we exclusively consider the final position, except for the initial segment where all positions are comprehensively evaluated, in alignment with the approach outlined in \cite{schlag2021linear,al2019character}. For our language modeling implementation, we rely on the publicly available code developed by \cite{schlag2021linear}.

\textbf{Model settings for WikiText-103}. In our experiments, we set the dimensions of keys, values, and queries to 128, while the training and evaluation context length is set to
256. Again, for $p$-LaT, we follow the same parameters settings, the difference is that we set 4 attention heads with low $p$ value ($p=1.5$), and the other 4 heads with high $p$ value ($p=2.5$).

As indicated in Table \ref{tab:result_wiki}, our $p$-LaT language model showcases superior performance, excelling in both test and valid perplexity metrics when contrasted with the softmax transformer language model \citep{xiong2021nystromformer}. These empirical outcomes, complemented by the success observed in diverse tasks, provide robust evidence of the substantial advantages offered by our $p$-LaT models.

\section{Related Work}
\label{sec:related}
\textbf{Transformers}. Transformers have left an indelible mark on various domains, with foundational work introducing the Transformer model \citep{vaswani2017attention}, and variations like BERT \citep{devlin2018bert} and GPT \citep{brown2020language} extending its applications in NLP. They've revolutionized transfer learning, with models like RoBERTa \citep{liu2019roberta} leading the way, while Vision Transformers (ViTs) \citep{dosovitskiy2020image} bring the Transformer architecture to computer vision. Efficiency-focused models like DistilBERT \citep{sanh2019distilbert} and MobileBERT \citep{sun2020mobilebert} maintain performance while reducing model size. In summary, Transformers continue to shape NLP, computer vision, and other domains, owing to their adaptability, transfer learning capacity, and versatility. However, it still has limitation. \cite{dong2021attention,wang2022anti} investigate Transformers in the image domain through the lens of Fourier spectrum, showing that self-attentions are low-pass filters, retaining only low-frequency, causing over-smoothed outputs. Our work is an orthogonal explanation of the previous work, providing a variational perspective of the low-frequency phenomenon and deriving
the novel $p$-LaT method to obtain non-low frequencies and improve the performances under implicit homophilic/heterophilic structures.

\textbf{$p$-Laplacian Regularization Framework}. The application of $p$-Laplacian regularization in image processing, particularly in the domain of denoising, has gained substantial attention. $p$-Laplacian regularization is a powerful tool for reducing noise while preserving essential image features. Prior studies such as \cite{pang2017graph} and \cite{elmoataz2015p} have explored the use of $p$-Laplacian regularization to enhance image denoising techniques. These works emphasize the effectiveness of $p$-Laplacian regularization in improving the quality of denoised images, making it a prominent approach in the field. Additionally, in the realm of machine learning, \cite{fu2022p} harnesses $p$-Laplacian regularization to address heterophily challenges within discrete graphs. Notably, \cite{calder2018game} and \cite{slepcev2019analysis} conduct thorough analyses of $p$-Laplacian regularization, particularly in semi-supervised and regression tasks, showcasing its competence when working with limited training data. It's noteworthy that the original $p$-Laplacian regularization is defined as follows:
\begin{equation*}
    J(\bm{u})=\frac{1}{p}\int_{\Omega^2}{k(x,y)\|\bm{u}(y)-\bm{u}(x)\|^pdydx} + \mu\|\bm{u}-\bm{x}\|^2 
\end{equation*}
Here, the first term quantifies signal variation over image patches or networks, while the second term enforces constraints to prevent excessive deviation from the input signal $\bm{x}$. The parameter $\mu \in (0,\infty)$ governs the balance between these constraints. In our work, we set $\mu=0$ to explore how traditional Transformers evolve with varying values of $p$.

\section{Conclusion}
In this study, we leveraged the $p$-Laplacian regularization framework to illuminate the inner workings of Transformers' representations. This exploration deepened our understanding and introduced pathways for performance enhancement. Under spectral analysis, our work unveiled $p$-Laplacian regularization as a versatile framework for self-attention mechanisms, guarantees in convergence and harnessing the concepts of both homophily and heterophily within Transformers to amplify their capabilities. We validated our approach through empirical assessments across large-scale applications, encompassing ImageNet object classification and WikiText-103 language modeling. However, it's important to acknowledge that our study does not address the robustness of our model to perturbed data. This presents an intriguing avenue for future research—exploring how the comprehensive $p$-Laplacian framework can bolster the robustness of Transformer models. We leave this idea for future investigation.

\bibliography{references}

\end{document}


%

%

\onecolumn
\aistatstitle{Supplementary Materials to ``$p$-Laplacian Transformer''}

\section{Further discussion}
\subsection{Transformers as Graph Learning}
In this section, we aim to establish a connection between the architecture of Transformers and the graph learning problem. Conceptualizing Transformers as graphs provides a valuable framework for understanding their inner workings and potential challenges.

When we view Transformers as graphs, each word or token within a sequence serves as a node in this graph. The relationships between these nodes are effectively represented by the edges connecting them. These edges in the graph signify the connections between tokens. In the self-attention mechanism, each token interacts with every other token, and the strength of these interactions is dictated by attention scores, which can be thought of as the weights assigned to the edges.

As information flows through the self-attention mechanism, it traverses the edges of this graph. Each node gathers and integrates information from its neighboring nodes based on the attention scores. This mechanism enables the model to focus on relevant tokens when processing a particular token, akin to how information propagates in a network. It's important to note that a Transformer model comprises multiple layers of self-attention, with each layer corresponding to a distinct level of the graph. The output of one layer serves as the input to the next, and information is hierarchically propagated through the graph, enabling the model to capture intricate dependencies in the data.

Crucially, the attention matrix in a Transformer can be seen as a structural representation of the graph. Thus, any modifications to the self-attention matrix effectively correspond to alterations in the graph's structure. This observation underscores the significance of considering the graph-based perspective when working with Transformers.

Moreover, it's noteworthy that, much like in traditional graph data structures, the issue of heterophily can manifest in Transformers. Heterophily denotes the preference for connections between dissimilar nodes, and it can be crucial in various tasks as we have listed some examples in the main text. Therefore, addressing the heterophily problem in Transformers becomes a pertinent concern, as it can significantly impact their performance in tasks that require attention to dissimilar tokens or nodes.

\section{Additional experiments}
This section provides extra empirical analysis to further demonstrate the benefits of $p$-LaT.

\subsection{Imagenet}
In the original DeIT model, they employed a multi-head attention mechanism with only 3 heads. In contrast, our $p$-LaT model utilizes 4 heads. To ensure a rigorous comparison, we conducted an additional experiment with the DeIT model, configuring it with four heads. Concurrently, we ran an extra experiment with the $p$-LaT model. In this experiment, we allocated two heads with $p=2$ and two heads with $p=2.5$, denoting this specific configuration as \textbf{$p$-LaT (5)}. Remarkably, our $p$-LaT model consistently outperforms the DeIT baseline across all cases, including the scenario with four heads, as indicated in Table \ref{tab:result_imagenet}. These results underscore the significant advantages of our proposed $p$-LaT model. To reiterate, the default \textbf{$p$-LaT} setting comprises two heads with $p=1.5$ and two heads with $p=2$.

\begin{table*}[ht]
\centering
    \caption{Top-1 and Top-5 accuracy (\%) of $p$-LaT vs DeiT on the ImageNet benchmark.}
    {\begin{tabular}{ccc}
         \hline
          Model & Top-1 Acc (\%)& Top-5 Acc (\%) \\  
         \hline 
         DeIT (3 heads) & 71.97 & 91.13\\
         \hline 
         DeIT (4 heads) & 72.43 & 91.2\\
         \hline 
         $p$-LaT& \textbf{72.78} & \textbf{91.31}\\
         \hline
         $p$-LaT (5)& \textbf{72.83} & \textbf{91.62}\\
         \hline
    \end{tabular}}
    \label{tab:result_imagenet}
\end{table*}

\subsection{WikiText-103}

In a similar context, we extended our experimentation to the WikiText-103 task. For this particular evaluation, we configured our model with 4 heads using $p=1.5$ and an additional 4 heads with $p=2.5$, and we denoted this specific setup as \textbf{$p$-LaT (4)}. It's essential to highlight that, by default, the Softmax Transformer employs 8 heads in its multi-head mechanism, while our standard $p$-LaT setting includes 4 heads with $p=1.5$ and 4 heads with $p=2$. Our $p$-LaT model consistently outperforms the Softmax Transformer baseline in all scenarios, as evidenced in Table \ref{tab:result_wiki}.

\begin{table*}[ht]
\centering
    \caption{Test and valid perplexity (Test PPL and Valid PPL) on WikiText-103 of $p$-LaT compared to the conventional softmax transformer.}
    {\begin{tabular}{ccc}
         \hline
          Model & Valid PPL&Test PPL \\  
         \hline 
         Softmax Transformer & 33.17 & 34.1\\
         \hline 
         $p$-LaT & \textbf{32.57} & \textbf{33.5}\\
         \hline
         $p$-LaT (4)& \textbf{32.44} & \textbf{33.46}\\
         \hline

    \end{tabular}}
    \label{tab:result_wiki}
\end{table*}